\documentclass[11pt]{article}

\usepackage[]{EMNLP2022}

\usepackage{times}
\usepackage{latexsym}

\usepackage{graphicx}
\usepackage{float}

\usepackage{makecell}
\usepackage{multirow}
\usepackage{tabularx}
\usepackage{xspace,mfirstuc,tabulary}
\usepackage{booktabs}
\usepackage{longtable}

\usepackage[ruled]{algorithm2e}

\usepackage[T1]{fontenc}

\usepackage[utf8]{inputenc}

\usepackage{microtype}

\usepackage{inconsolata}

\title{DFM: Dialogue Foundation Model for Universal Large-Scale Dialogue-Oriented Task Learning} 

\author{Zhi Chen$^{1}$, Jiabao Ji$^{1}$, Lu Chen$^{1}$, Yuncong Liu$^{1}$, Da Ma$^{1}$, Bei Chen$^{2}$,
\\ \textbf{Mengyue Wu}$^{1}$, \textbf{Su Zhu}$^{3}$, \textbf{Xin Dong}$^{3}$, \textbf{Fujiang Ge}$^{3}$, \textbf{Qingliang Miao}$^{3}$, \textbf{Jian-Guang Lou}$^{2}$ \and \textbf{Kai Yu}$^{1}$ \\
        $^1$X-LANCE Lab, Department of Computer Science and Engineering \\ 
        MoE Key Lab of Artificial Intelligence, AI Institute, Shanghai Jiao Tong University\\
    State Key Lab of Media Convergence Production Technology and Systems, Beijing, China\\
    $^2$Microsoft Research Asia \\
    $^3$AISpeech Co., Ltd., Suzhou, China \\
        }

\begin{document}
\maketitle

\begin{abstract}
Building a universal conversational agent has been a long-standing goal of the dialogue research community. Most previous works only focus on a small set of dialogue tasks. In this work, we aim to build a unified dialogue foundation model (DFM) which can be used to solve massive diverse dialogue tasks.  To achieve this goal, a large-scale well-annotated dialogue dataset with rich task diversity (DialogZoo) is collected. We introduce a framework to unify all dialogue tasks and propose novel auxiliary self-supervised tasks to achieve stable training of DFM on the highly diverse large scale DialogZoo corpus. Experiments show that, compared with models of the same size, DFM can achieve state-of-the-art or competitive performance on very rich cross-domain downstream dialogue tasks. This demonstrates that DFM largely extends the ability of unified dialogue pre-trained model. 
\end{abstract}

\section{Introduction}
\label{sec:intro}

Nowadays, dialogue systems are widely applied in various scenarios, such as intelligent personal assistants, customer service centers, smart speakers and so on. 
These applications have extensively promoted dialogue researches, with various dialogue tasks being proposed, e.g. intent classification, dialogue state tracking, conversational question answering, chit-chat, etc. 
For each task, multiple dialogue datasets are collected and carefully annotated. Consequently, we have witnessed rapid developments of dialogue models and methods \cite{gao2019neural}. 
However, most of them are targeted at a single task, sometimes even a single dataset. 
This phenomenon has limited the possibility to transfer knowledge between tasks and datasets, which hence undermined the development of  unified and well-generalized dialogue systems.

Thanks to the development of pre-trained language models (PLMs) \cite{qiu2020pre}, more and more research focus on building unified models for different natural language processing tasks. In the dialogue area, there are also some attempts to build unified dialogue models. However, previous works only focus on a small subset of dialogue tasks, e.g. dialogue understanding \cite{chen2022unidu}, and task-oriented dialogue \cite{su2021multi}. Building unified dialogue models for massive diverse dialogue tasks has been a long-standing goal in the community.  

\begin{figure}[t]
\centering
\includegraphics[width=0.48\textwidth]{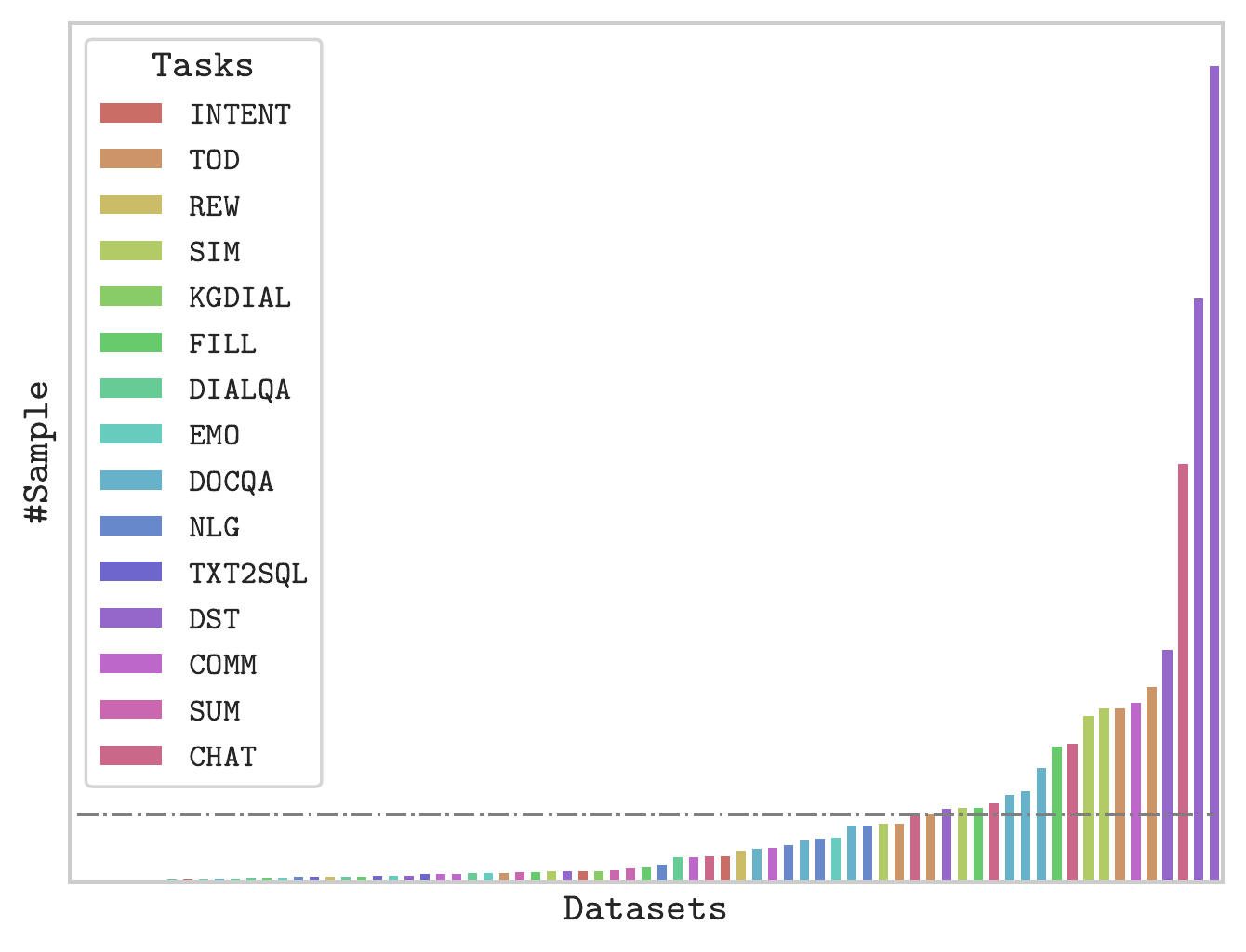} 
\caption{The number distribution of the training samples on 73 dialog corpora, which span 15 supervised dialogue-oriented tasks.}
\label{fig:statis}
\vspace{-0.2cm}
\end{figure}

In this paper, we aim to train a unified \textbf{D}ialogue \textbf{F}oundation \textbf{M}odel (DFM) for a plethora of dialogue tasks. Such a model should be entitled to two types of skills: 1) Superb dialogue understanding capability, no matter the dialogue is a single utterance or long conversation with multi-turns. It should correctly  extract useful information from the dialogue, e.g. entities, dialogue state, user's intent and emotion. 2) Being able to leverage external knowledge, which can be plain text, semi-structured description, and structured knowledge graph (KG) or databases. These skills can not be achieved by traditional self-supervised training. 

Therefore, we collect a large-scale dialogue dataset, {\bf DialogZoo}, with rich diversified tasks from publicly available datasets. The dataset consists of 15 dialogue tasks, involving as many as 73 datasets. These tasks include \textit{intent detection, slot filling, dialogue state tracking, dialogue emotion recognition, KG-guided dialogue, Text-to-SQL, etc}. 
Both the input and output of these tasks are very diverse. 
The distribution of training samples supporting each task is shown in Figure~\ref{fig:statis}. 

In order to train a unified model, the primary prerequisite is to translate data of different tasks into a unified form.  
Motivated by instruction tuning \cite{wei2021finetuned} and prompt learning \cite{liu2021pre}, we restructure the input of each dialogue task into three parts: dialogue content , external knowledge, and task description, and normalize all data into text-to-text format. 
With this unified format, we can directly train the dialogue model initialized from a typical seq2seq PLM, e.g BART~\cite{lewis2020bart} and T5~\cite{raffel2020exploring}. However, our preliminary experiments suggest that straightforward supervised training is not stable for these extremely diverse dialog tasks.
Therefore, we further propose two dialogue-oriented self-supervised tasks, and finally use the mixture of supervised and self-supervised methods to successfully train our {\bf Dialogue Foundation Model} (DFM). 
The supervised learning paradigm enables the model to learn task-specific skills, while the self-supervised tasks are tailored for dialogue recovery skill. 
Upon the inclusion of rich diverse datasets and a hybrid training procedure, DFM can be readily used in various downstream dialogue tasks. 
Extensive experimental results indicate that DFM not only improves seq2seq generation performance, but also enhances the dialogue representation ability. 

The contribution of this paper comes three folds:
\begin{itemize}
    \item We  collect a large-scale dialogue dataset (DialogZoo) across 15 dialogue tasks and 73 dialogue datasets, and design a unified text-to-text input/output format for all tasks. It provides a new benchmark which can facilitate researching on unified dialogue models. To the best of our knowledge, this is by far the largest and richest form-unified dialogue dataset supporting as many as 15 different dialogue tasks. We will make it publicly available.
    \item We introduce two dialogue-oriented self-supervised tasks as complementary tasks with other supervised tasks. A unified dialogue foundation model (DFM) can then be stably trained with multi-task learning on the DialogZoo dataset. 
    \item Extensive experimental results show that our pre-trained DFM outperforms the state-of-the-art models on three aspects including dialogue representation, knowledge distillation and generation.
\end{itemize}

\section{Related Work}

\subsection{Pre-trained Dialogue Models}
Pre-trained dialogue models usually refer to pre-trained language models (PLMs) designed for dialogue tasks. 
Recent years have witnessed remarkable progress in PLMs. Most of them fall into three categories. The first is training language models on large-scale dialogue corpus using typical self-supervised objectives, i.e. mask language modeling (MLM) \cite{devlin2019bert} and autoregressive generation \cite{brown2020language}. Representative works that fall into this category include Meena \cite{adiwardana2020towards}, Blender \cite{roller2020recipes}, and DialoGPT \cite{zhang2020dialogpt},  which mainly utilize large-scale open-domain dialogues as the training corpus, and TOD-BERT \cite{wu-etal-2020-tod} 
which use task-oriented dialogues as the training corpus. The utilization of dialogue corpus significantly boosts the dialogue understanding and generation ability of language models. The second category is training language models on well-labeled dialogues using task-specific supervised objectives. Most works that fall into this category focus on the task-oriented dialogue area, such as SOLOIST \cite{peng2021soloist}, PPTOD~\cite{su2021multi}, and GALAXY \cite{he2022galaxy}. The usage of dialogue annotations not only improves the performance of related downstream tasks but also improves the transferability of dialogue models in few-shot scenarios. The third is to design dialogue-oriented self-supervised training objectives \cite{bao2020plato,xu2021dialogue}. These works try to minimize the gap between self-supervised tasks and downstream dialogue tasks.

Different from the above works, we collect a large scale well-annotated dialogue corpus of very diverse dialogue tasks and formulate all of them into a unified text-to-text format. With this supervised corpus, we further propose two dialogue-oriented self-supervised tasks to jointly pre-train our unified dialogue model.

\begin{table*}
\centering
\small
\setlength{\tabcolsep}{0.8mm}{
\begin{tabular}{c|c|c|c|c}
\hline
\textbf{DOTs} & \textbf{Dialog History} & \textbf{External Knowledge} & \textbf{Output Format} & \textbf{\#Dataset} \\
\hline
Dialogue Rewrite (R{\small{EW}}) & Multiple Turns & None &  Single Utterance & 3 \\
Natural Language Generation (N{\small{LG}}) & None & Semi-structured Description & Single Utterance & 5 \\
Dialogue Summary (S{\small{UM}}) & Multiple Turns & None  & Summary Text & 3 \\
Slot Filling (F{\small{ILL}}) & Single Turn &  Semi-structured Description & Logical Form & 7  \\
Intent Detection (I{\small{NTENT}}) & Single Turn & Semi-structured Description & Classification Name & 4 \\
Dialogue State Tracking (D{\small{ST}}) & Multiple Turns & Semi-structured Description & Logical Form & 6 \\
Commonsense QA (C{\small{OMM}}) & Single Turn & Unstructured Text & Fact of Phrase & 5 \\
Emotion Detection (E{\small{MO}}) & Single Turn & Semi-structured Description & Classification Name & 5 \\
Document QA (D{\small{OCQA}}) & Multiple Turns & Unstructured Text & Fact of Phrase & 7 \\
Dialogue QA (D{\small{IALQA}}) & Multiple Turns & Unstructured Text & Fact of Phrase & 6 \\
Chitchat (C{\small{HAT}}) & Multiple Turns & Unstructured Text & Single Utterance & 5 \\
Knowledge-graph Dialogue (K{\small{GDIAL}}) & Multiple Turns & Structured Knowledge & Single Utterance & 2 \\
Text-to-SQL (T{\small{XT2SQL}}) & Multiple Turns & Structured Knowledge & Logical Form & 3 \\
User Simulator (S{\small{IM}}) & Multiple Turns & Semi-structured Description & Single Utterance & 6 \\ 
Task-oriented Dialogue (T{\small{OD}}) & Multiple Turns & Semi-structured Description & Single Utterance & 6 \\ 
\hline
\end{tabular}
}
\caption{\label{tab:tasks}
The detailed introduction of 15 dialogue-oriented tasks (DOTs) under the unified generative framework. ``\#Dataset'' means the number of datasets used in corresponding task.
}
\vspace{-0.2cm}
\end{table*}

\subsection{Multi-task Learning for Pre-trained Language Models}

Recently, multi-task learning of pre-trained language models has attracted spotlight research interests. \citet{aghajanyan2021muppet} proposes the concept of pre-finetuning, which is massive multi-task learning on 50 tasks. They show that pre-finetuning consistently improves performance for pre-trained models on various classification and generation tasks. FLAT \cite{wei2021finetuned} and T0 \cite{sanh2021multitask}  scale to more tasks (i.e. 62 and 171 respectively) for multi-task learning, and translate all samples to text-to-text formats based on task instructions. The most recent proposed ExT5 \cite{aribandi2021ext5} is a model pre-trained using multi-task learning of self-supervised span denoising and supervised datasets.

In the dialogue research field, there are also a few works on multi-task learning of pre-trained language models. \citet{su2021multi} proposes a multi-task pre-training framework that allows a unified dialogue model to solve different sub-tasks in task-oriented dialogue systems. \citet{chen2022unidu} investigates different multi-task training strategies for a set of dialogue understanding tasks. 
Our work distinguishes from the above by not limiting to a subset of dialogue tasks. We explore large-scale multi-task learning of enormous yet diverse supervised dialogue tasks as well as two newly proposed dialogue-oriented self-supervised tasks.

\section{DialogZoo}
As we know, dialogue always happens on specific scenarios, which is also called dialogue content~\cite{deriu2021survey}. In existing dialogue tasks, the dialogue content mainly contains dialogue history and the corresponding grounded knowledge. For example, the knowledge in text-to-SQL is schema of database and the knowledge in task-oriented dialogue task is dialogue ontology designed by domain experts.  However, dialogue corpus with paired dialogue and knowledge is hard to collect from daily conversations, like Reddit. Instead, we choose to collect the existing well-annotated dialogue corpora to train a unified dialog model.

\subsection{Dialogue-Oriented Tasks}
All the collected corpora in this paper are dialogue-oriented. Our main principle is to cover as many diverse dialogue formats as possible. We finally collect 73 dialogue datasets, named {\bf DialogZoo}, which span 15 different dialogue-oriented tasks. DialogZoo contains various dialogue types: single-turn utterance or question, sequential questions, multi-turn chitchat and task-oriented dialogue. The formats of external knowledge in DialogZoo are also diverse. They include world knowledge, unstructured text, semi-structured description and structured knowledge. Note that world knowledge means the knowledge is not given, which needs the dialogue model to reason, like dialogue summary task. Due to rich diversity of dialogue-oriented tasks, DialogZoo contains a wide variety of outputs formats including classification names (like in intent detection), plain text utterance (like in dialogue rewrite), logical forms (like in text-to-SQL) and so on. The details are shown in Table~\ref{tab:tasks}.



\subsection{Knowledge-Grounded Dialogue}
One intrinsic characteristic of dialogue is that the dialogue system is knowledge-intensive. Dialogue understanding and generation heavily rely on the given external knowledge. For example, all available intents need be explicitly designed by experts before performing dialogue intent detection. Hence, a unified dialogue model needs to integrate external knowledge in addition to dialogue content.


\begin{figure}[t]
\centering
\includegraphics[width=0.48\textwidth]{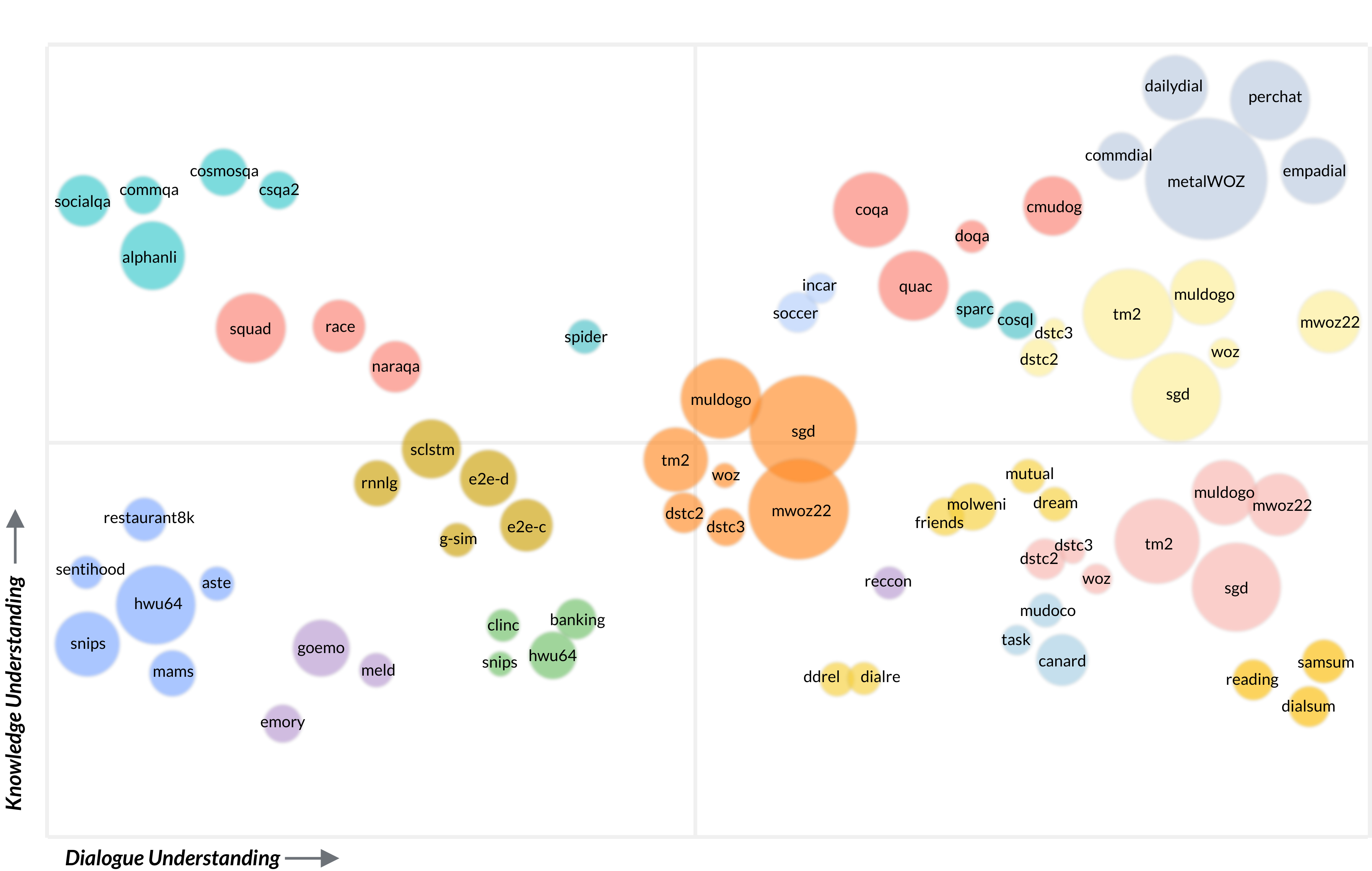} 
\caption{All of the collected dialogue datasets in DialogZoo. The upper right means that the dialog tasks need stronger understanding ability on both dialogue and knowledge levels.}
\label{fig:dataset-2dim}
\vspace{-0.2cm}
\end{figure}

In this paper, the unified dialogue model is expected to have rich abilities of \textbf{dialogue understanding} and \textbf{knowledge understanding}. Hence, we differentiate the datasets of DialogZoo according to these two dimensions, as shown in Figure~\ref{fig:dataset-2dim}. The x-axis of dialog understanding measures the complexity of dialogue content. The dialogue turns become longer from left to right. The y-axis of external knowledge measures the difficulty of extracting useful information. At the bottom of the quadrant diagram, external knowledge is usually well-organised and structured. At the top, external knowledge may just be document text, which contains large amounts of redundant information and increases the difficulty of understanding than well-labelled structured knowledge. Specially, in this paper, we put the chitchat task at the top, where the external knowledge is not given. The chitchat task requests the dialogue agent to remember the related world knowledge by itself. Note that the size of circles in Figure~\ref{fig:dataset-2dim} is proportional to the scale of annotated samples.

\begin{figure*}[t]
\centering
\includegraphics[width=0.85\textwidth]{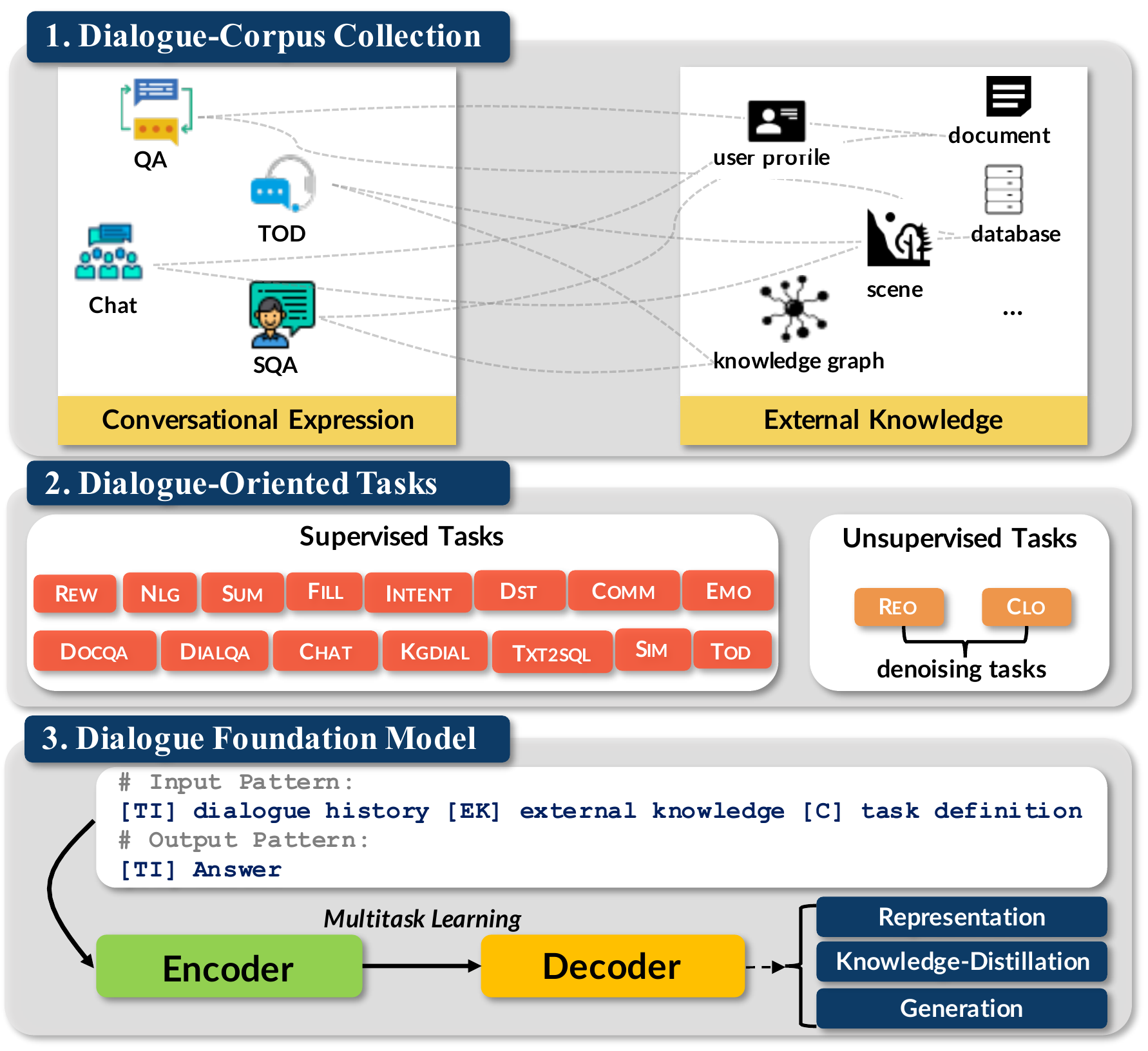} 
\caption{The 3-step construction pipeline for dialogue foundation model (DFM): 1) dialogue-corpus collection: collecting the popular dialogue datasets; 2) dialogue-oriented tasks: designing two unsupervised denoising tasks; 3) unified DOT modeling: training the unified DFM with multitask learning method.}
\vspace{-0.4cm}
\label{fig:dialogzoo}
\end{figure*}

\section{Dialogue Foundation Model}
In this section, we first design a framework to formulate all the dialogue-oriented tasks into text-to-text format, named dialogue foundation model (DFM). To enhance representation ability of DFM, we further design two self-supervised tasks (named R{\small EO} and C{\small LO}). Last but not least, we introduce an effective training strategy for large-scale highly-diverse dialogue-oriented tasks, where the training samples of different tasks are unbalanced.

\subsection{Unified Generative Framework}
\label{sec:ugf}
As shown in Figure~\ref{fig:dialogzoo}, dialogue corpora always connect with the diverse external knowledge. For example, the external knowledge of QA data is normally sourced from document and the task-oriented dialog (TOD) is associated with a database, which is defined by the expert-designed dialog ontology. 
In DialogZoo, all dialogue types are included in the dialogue history, e.g., single turn QA, sequential QA (SQA), task-oriented dialogue (TOD) and chitchat (Chat). The types of corresponding external knowledge are also diverse, including unstructured document, semi-structured scene description, structured database etc.

Given this circumstance, we split the input of dialogue-oriented tasks (DOTs) into three components: \textit{dialogue history}, \textit{connected external knowledge} and \textit{task definition}. The output format under the unified generative framework is dependent on the task definition. The unified generative template is shown in third part of Figure~\ref{fig:dialogzoo},
\noindent where ``[TI]'' is the special token of dialogue-oriented tasks, e.g., ``[rew]'' represents the task identification of dialogue rewrite (R{\small EW}) task. The dialogue history and the corresponding external knowledge have various input formats across 15 dialogue-oriented tasks. 
The input dialogue content consists of three categories: single turn, multiple turns and ``None'', where ``None'' dialogue content exists in N{\small LG} task, whose input is the logical form. 
The external knowledge span four categories: unstructured text, semi-structured description, structured knowledge and ``None''. 
Here, the special case ``None'' exists in the general dialogue-oriented tasks, e.g, dialogue summary. 
The task definition is the custom description with only one sentence. The output formats are also diverse, which consist of unstructured text (e.g., in I{\small NTENT} and S{\small UM}), single utterance (e.g., in R{\small EW} and N{\small LG}), structured knowledge (e.g., in T{\small XT2SQL}). Note that the output of these supervised tasks never involves complete dialogue content. The detailed introduction of the input and output on all DOTs are shown in Table~\ref{tab:tasks}, with specific instances shown in Appendix.

\subsection{Self-supervised Tasks}
As mentioned in Section~\ref{sec:ugf}, the complete dialogue content is never covered by the output of the existing supervised tasks in DialogZoo. 
Therefore, the decoder of DFM is limited to learn information at turn-level, which may restrict its dialog-level content representation capability.
To alleviate this problem, we design two self-supervised dialogue tasks: R{\small EO} and C{\small LO}, which are both dialogue-level denoising tasks. 

\noindent \textbf{R{\small EO} Task} Under the DFM framework, the dialogue content in R{\small EO} is the aggregation of the multiple-turn dialogue data in all  supervised corpora. 
We randomly permute the order of mult-turn dialogue at turn level. The R{\small EO} task aims to recover the original order of the permuted dialogue. The input external knowledge of R{\small EO} is ``None''. 

\noindent \textbf{C{\small LO} Task} Similarly, the dialogue content in C{\small LO} is the aggregation of single-turn and multi-turn dialogue data in all supervised corpora. 
We leverage the entity extraction tool (spaCy)~\footnote{\url{https://github.com/explosion/spaCy}} to extract all meaningful entities in the dialogue content. The extracted entities are replaced by masking tokens in the dialogue content and the set of extracted entities are permuted, working as external knowledge.  
The C{\small LO} task aims to recover the complete dialogue content given the entities from external knowledge and the masked dialogue content.

\subsection{Task-iterative Training Strategy}
As shown in Figure~\ref{fig:statis}, the distribution of training samples is extremely unbalanced. Our preliminary results show that direct mix-up training makes the model converge to the dominant tasks, e.g., dialogue state tracking (D{\small ST}). 

Therefore, we apply a task-iterative training strategy, where a hyperparameter is used to control the task iteration step (e.g., 512 samples in our experiments). In the multi-task training process, we iterate the dialogue-oriented tasks (including supervised and self-supervised tasks) at each 512 samples. The epoch ends until the data of the largest task are all swept. The rest tasks may be trained more than one time. This task-iterative training strategy can guarantee that every DOT is trained by the same steps at each epoch, purposefully applied to achieve comparable results on all DOTs.
The detailed training strategies are shown in Algorithm~\ref{algo:tits} of Appendix.
After multi-task pre-training, DFM can be directly used for tasks in DialogZoo, or further fine-tuned on specific downstream datasets.

\begin{table*}[h!]
\centering
\small
\newcolumntype{s}{>{\columncolor[HTML]{DCDCDC}} c}
\setlength{\tabcolsep}{0.8mm}{
\begin{tabular}{c | c c c c c c} 
 \bottomrule
  \hline
 \multirow{2}{10em}{\centering DialoGLUE} & \multicolumn{6}{c}{PLMs} \\ 
 & ConvBERT & ConvBERT$^*$ & BART$_{\rm base}$ & DFM$_{\rm bart}$ & T5$_{\rm base}$ & DFM$_{\rm t5}$ \\
 \hline
 \multicolumn{7}{l}{\textbf{Intent Detection}} \\
  B{\small ANKING77}~(ACC.) & 92.89 & 92.08 & 93.15 & \textbf{93.32} & 92.31 & 93.02 \\
  C{\small LINC150}~(ACC.) & 97.04 & 95.58 & 96.20 & \textbf{97.33} & 96.49 &  97.31\\
  H{\small WU64}~(ACC.) & 91.82 & 90.89 & 91.17 & \textbf{94.42} & 91.26 & 93.16\\
\hline
 \multicolumn{7}{l}{\textbf{Slot Filling}} \\
 R{\small ESTUARANT8K}~(F1) & 96.04 & 95.29 & 92.29 & 93.89 & 95.98 &  \textbf{96.07} \\
 D{\small STC8}~(F1) & 89.45 & 86.71 & 86.77 & 88.53 & 90.56 & \textbf{90.76}  \\
 \hline
 \multicolumn{7}{l}{\textbf{Semantic Parsing}} \\
 T{\small OP}~(EM) & 81.43 & 80.80 & 80.71 & 81.24 & 81.81 &  \textbf{82.04} \\
 \hline
 \multicolumn{7}{l}{\textbf{Dialog State Tracking}} \\
 M{\small ULTIWOZ2.1}~(JGA) & 58.11 & 56.21 & 56.57 & 59.19 & 57.21 & \textbf{60.58}  \\
 \hline
 Average Score & 86.68 & 85.37 & 85.27 & 86.85 & 86.73 & \textbf{87.56} \\
 \hline
\end{tabular}
}
\caption{The representation ability of six different pre-trained language models on DialoGLUE benchmark. All metrics are the higher the better.}
\vspace{-0.4cm}
\label{tab:rep}
\end{table*}

\section{Experiments}
We train the unified dialogue foundation model (DFM) on DialogZoo with the 15 existing supervised dialog-oriented tasks and the 2 newly proposed self-supervised tasks. The DFM can be used in three primary ways: \textit{\textbf{representation}}, \textit{\textbf{knowledge-distillation}} and \textit{\textbf{generation}}. 
Representation means that DFM can be directly initialized as a dialogue encoding model and then be fine-tuned in downstream tasks. Knowledge-distillation indicates that DFM can serve as a dialogue understanding or information/knowledge extraction model. Generation means that it is also a dialogue response model, which can generate fluent and reasonable text  expression. Compared with previous pre-trained dialogue models, DFM is the only one that can achieve strong performance on all the three aspects. 



\subsection{Implementation}
\begin{figure}[t]
\centering
\includegraphics[width=0.5\textwidth]{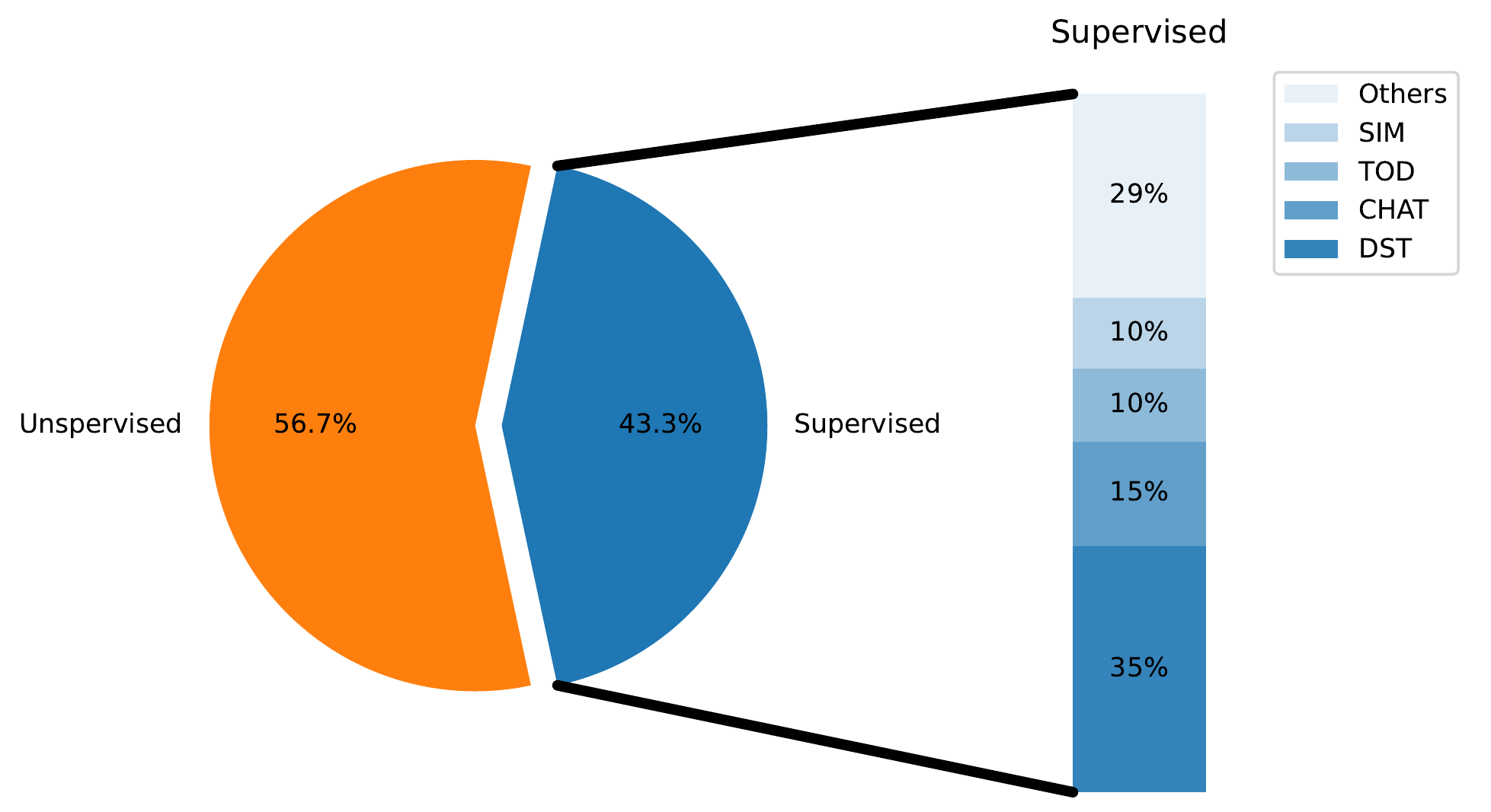} 
\caption{The proportion of the training samples on unsupervised tasks and supervised tasks. The top 4 supervised tasks that have more training samples are D{\small ST}, C{\small HAT}, T{\small OD} and S{\small IM}.}
\vspace{-0.3cm}
\label{fig:rate}
\end{figure}

We respectively use BART-Base~\cite{lewis2020bart} and T5-Base~\cite{raffel2020exploring} as the backbones for DFM. All the Transformer-based pre-trained language models are loaded from HuggingFace library~\cite{wolf2019huggingface}. We train the DFM models on 8 V100 GPUs with 32G memory. The batch size is 256 with gradient accumulation strategy (updated per 8 steps). The max token length of input and output sets as 350. The learning rate is 1e-5 with AdamW optimizer. DFM models are trained with 200K steps. We finally involved 10,661,579 training samples, including supervised instances and self-supervised instances, which extract from only training sets of collected datasets. The detailed proportion of training samples is illustrated in Figure~\ref{fig:rate}.

\subsection{Usage 1: Representation}
We conduct all the representation experiments on the standard DialoGLUE benchmark~\cite{mehri2020dialoglue}, which consists of four dialogue-oriented tasks: slot filling (F{\small ILL}), intent detection (I{\small NTENT}), semantic parsing and dialogue state tracking (D{\small ST}). TOP~\cite{gupta2018semantic} is semantic parsing task in DialoGLUE, which is unseen in DialogZoo. The strong baseline model ConvBERT~\cite{mehri2020dialoglue} is pre-trained on over 70 million Reddit data with MLM method~\cite{devlin2019bert}. To fairly compare with our proposed DFM, we also pre-train the comparable ConvBERT$^*$ with our collected dialogue corpora. 
The 4.6 million DialogZoo data used to train ConvBERT$^*$ is substantially less than the original corpus in ConvBERT.
We rerun ConvBERT and ConvBERT$^*$ on DialoGLUE with their released source code. All the test results are based on the checkpoints that achieve the best evaluation performance on development set. As shown in Table~\ref{tab:rep}, we can see that the representation ability of ConvBERT$^*$ is absolutely worse than the ConvBERT. The main reason is that the scale of pre-training corpus is less in an order of magnitude, which is the main factor affecting unsupervised learning performance. 

We directly replace the ConvBERT model with our DFM, where we do not adjust any hyper-parameters during the fine-tuning process. The designed models for downstream tasks in DialoGLUE are also unchanged, e.g, DST used Trippy model~\cite{heck2020trippy}. DFM$_{\rm bart}$ and DFM$_{\rm t5}$ means that the backbones of DFM are BART-Base and T5-Base, respectively. Compared with original backbones, two DFMs achieved better performance on all DialoGLUE tasks. DFM$_{\rm bart}$ enjoys the best performance on intent detection tasks. However, there is a large performance decrease in slot filling tasks compared to ConvBERT. DFM$_{\rm t5}$ surpasses ConvBERT on all tasks, where the average score gets absolute 0.88 points gain. It indicates that the multitask learning method is another way to enhance the representation ability of the pre-trained model with the same scale corpus. Compared with DFM$_{bart}$, DFM$_{t5}$ has obvious performance gain in DialoGLUE.

\begin{figure}[t]
\centering
\includegraphics[width=0.48\textwidth]{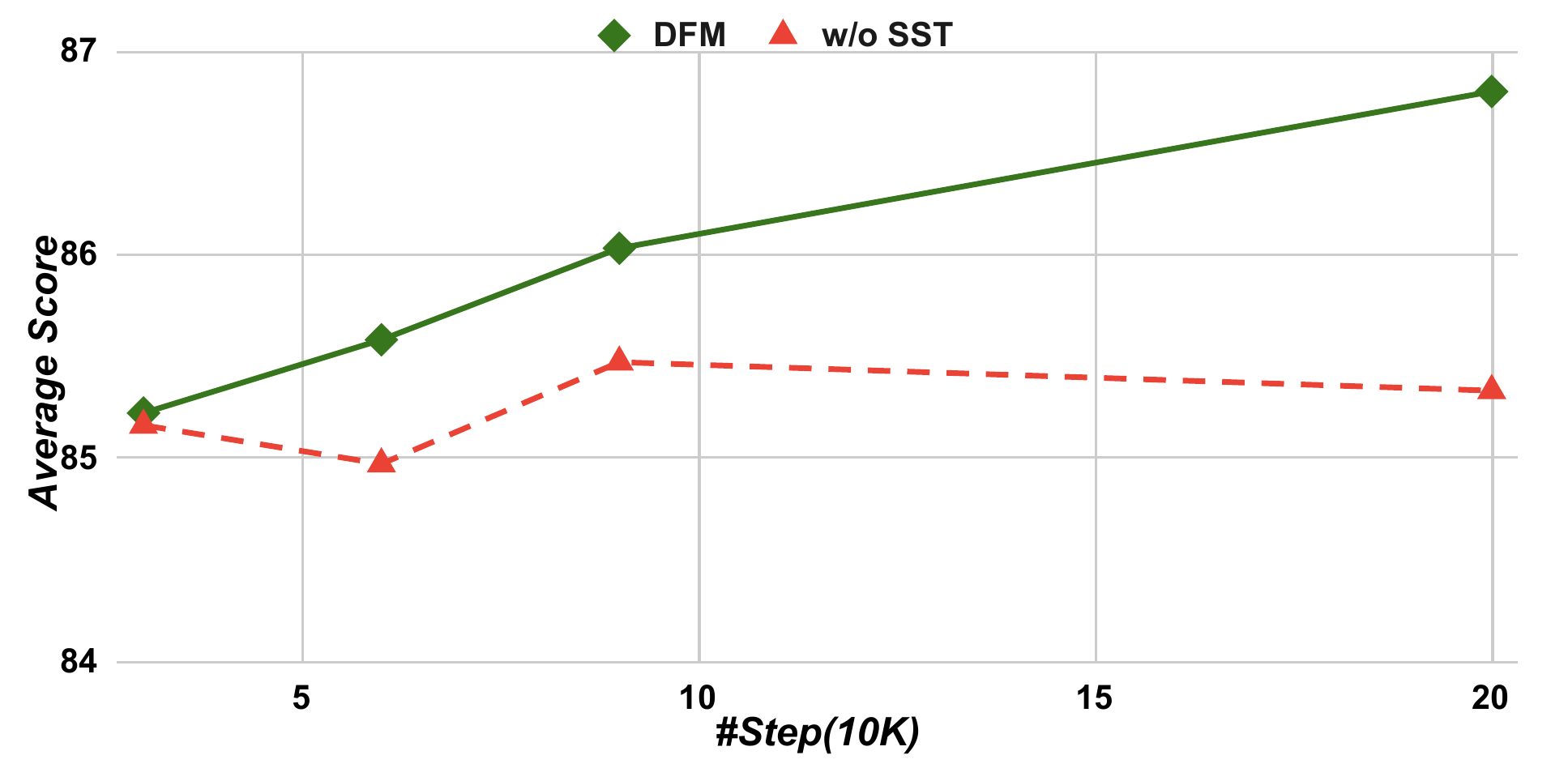} 
\caption{The average scores of DFMs with and without two self-supervised denoising tasks.}
\label{fig:abla}
\end{figure}

\noindent \textbf{Ablation Study} We remove two self-supervised tasks in DialogZoo to investigate their effects on representation ability. As shown in Figure~\ref{fig:abla}, ``w/o SST'' means the unified model is trained on only supervised dialogue-oriented tasks with backbone of BART-Base. We can see that the average scores on DialoGLUE are unstable during the training process. DFM with two unsupervised tasks can achieve stable growth on average score.

\begin{table}[t]
\centering
\small
\newcolumntype{s}{>{\columncolor[HTML]{DCDCDC}} c}
\setlength{\tabcolsep}{0.8mm}{
\begin{tabular}{c | c c c} 
 \bottomrule
  \hline
 \multirow{2}{8em}{\centering Knowledge Distillation} & \multicolumn{3}{c}{Methods} \\ 
 & SOTA & DFM$_{\rm t5}$ & w/ FT  \\
 \hline
 \multicolumn{4}{l}{\textbf{Intent Detection}} \\
  B{\small ANKING77}~(ACC.) & \textbf{93.86}~(\citeyear{su2021multi}) & 92.82 & 93.63 \\
  C{\small LINC150}~(ACC.) & 97.16~(\citeyear{casanueva2020efficient}) & \textbf{100} & \textbf{100} \\
  H{\small WU64}~(ACC.) & 92.94~(\citeyear{mehri2020dialoglue}) & \textbf{98.70} & 98.42 \\
\hline
 \multicolumn{4}{l}{\textbf{Slot Filling}} \\
 R{\small ESTUARANT8K}~(F1) & 98.00~(\citeyear{peng2021soloist}) & 95.95 & \textbf{99.28} \\
 D{\small STC8}~(F1)$^\dagger$ & 89.45~(\citeyear{mehri2020dialoglue}) & 47.85 & \textbf{89.69}  \\
 \hline
 \multicolumn{4}{l}{\textbf{Semantic Parsing}} \\
 T{\small OP}~(EM)$^\ddagger$ & 86.67~(\citeyear{rongali2020don}) & / & \textbf{87.10} \\
 \hline
 \multicolumn{4}{l}{\textbf{Dialogue State Tracking}} \\
 M{\small ULTIWOZ2.2}~(JGA) & 57.70~(\citeyear{sun2022tracking}) & 58.52 & \textbf{59.20}  \\
 \hline
\end{tabular}
}
\caption{The performance on knowledge distillation tasks. $^\dagger$ means the corresponding dataset on slot filling task is unseen in DialogZoo. $^\ddagger$ means the task is unseen in DialogZoo.}
\vspace{-0.2cm}
\label{tab:kd}
\end{table}

\subsection{Usage 2: Knowledge-Distillation}
In addition to using DFM as representation model, we can also leverage DFM as knowledge distillation model to directly parse the structured knowledge in a generative way. Note that knowledge distillation here means the outputs of DOTs are structured logical form, which is well-organized knowledge. Due to the advance representation performance of DFM$_{t5}$ and content limitation, we just evaluate pre-trained DFM$_{t5}$ model on the task of DialoGLUE. Note that we replace M{\small ULTIWOZ2.1} with M{\small ULTIWOZ2.2}, which is the clean-annotation version used in DialogZoo. 
As shown in Table~\ref{tab:kd}, DFM$_{\rm t5}$ can achieve new state-of-the-arts on three knowledge distillation tasks. On C{\small LINC15}, DFM$_{\rm t5}$ even reaches 100\% accuracy. ``w/ FT'' means that DFM$_{\rm t5}$ continues to be fine-tuned on the corresponding tasks. The fine-tuned DFM$_{\rm t5}$ achieve another four SOTA performance on the distillation tasks. Note that D{\small STC8} on slot filling task is unseen on DialogZoo, which is a zero-shot setting. DFM$_{\rm t5}$ can get 47.85 F1 score. It indicates that DFM has strong generalization capability. 
On DST task, DFM can also surpass the well-designed SOTA model.
For DialogZoo, T{\small OP} is an unseen task, which is firstly formulated as a generation task and then fine-tuned on DFM$_{\rm t5}$. For the new task, our DFM$_{\rm t5}$ can still surpass SOTA with absolute 0.33 points. 
To further validate the efficiency of DFM, we leverage our DFM on more challenging T{\small XT2SQL} benchmarks.

\begin{table}[t]
\centering
\small
\newcolumntype{s}{>{\columncolor[HTML]{DCDCDC}} c}
\setlength{\tabcolsep}{0.8mm}{
\begin{tabular}{c | c c c} 
 \bottomrule
  \hline
 \multirow{2}{8em}{\centering Datasets} & \multicolumn{3}{c}{Methods} \\ 
 & T5$_{\rm base}$ & U{\small NISKG}$_{\rm base}$ & DFM$_{\rm t5}$  \\
 \hline
 \multicolumn{4}{l}{\textbf{With Beam Search Decoding}} \\
 Spider~(EM) & 57.20 & 59.86 & \textbf{59.96} \\
 CoSQL~(QEM) & 42.30 & 45.68 & \textbf{46.57}  \\
 \hline
 \multicolumn{4}{l}{\textbf{With PICARD Decoding}} \\
 Spider~(EM) & 65.80 & / & \textbf{68.47} \\
 CoSQL~(QEM) & 47.90 & / & \textbf{52.04} \\
 \hline
\end{tabular}
}
\caption{The performance on development sets of text-to-SQL benchmarks: Spider and CoSQL, where EM and QEM are exact match accuracy and question match accuracy.}
\vspace{-0.3cm}
\label{tab:txt2sql}
\end{table}

\noindent \textbf{Application in T{\small XT2SQL}}
We conduct the experiments on Spider and CoSQL benchmarks. The database schema on development set are unseen in training data, which is challenging zero-shot text-to-SQL setup. Compared with single-turn Spider benchmark, CoSQL is a sequential multi-turn text-to-SQL benchmark. There are lots of co-reference and information ellipse in dialogue content. To fairly compare to DFM$_{\rm t5}$, we leverage UNISKG$_{\rm base}$~\cite{xie2022unifiedskg} as the baseline, which is a special pre-trained model for structured inputs with T5-base as backbone. As shown in Table~\ref{tab:txt2sql}, DFM$_{\rm t5}$ has comparable performance on Spider with UNISKG$_{\rm base}$. However, on difficult CoSQL benchmark, our DFM$_{\rm t5}$ has a distinct advantage compared to vanilla T5$_{\rm base}$ and UNISKG$_{\rm base}$ with beam search decoding method. DFM$_{\rm t5}$ still maintains the advantages on these two benchmarks with PICARD decoding~\cite{scholak2021picard}, which is a grammar-constrained decoding method.

\begin{table}[t]
\centering
\small
\newcolumntype{s}{>{\columncolor[HTML]{DCDCDC}} c}
\setlength{\tabcolsep}{0.8mm}{
\begin{tabular}{c | c c c} 
 \bottomrule
  \hline
 \multirow{2}{8em}{\centering Generation} & \multicolumn{3}{c}{Methods} \\ 
 & SOTA & DFM$_{\rm t5}$ & w/ FT  \\
 \hline
 \multicolumn{4}{l}{\textbf{Chitchat}} \\
  P{\small ERSONALCHAT}~(R-L) & 15.22~(\citeyear{tay2021synthesizer}) & 23.53 & \textbf{25.79} \\
\hline
 \multicolumn{4}{l}{\textbf{Question Answering}} \\
 C{\small OQA}~(ACC.) & \textbf{90.70}~(\citeyear{DBLP:journals/corr/abs-1909-10772}) & 65.20 & 80.30  \\
 \hline
 \multicolumn{4}{l}{\textbf{Dialog Summary}} \\
 S{\small AMSUM}~(R-L) & 45.20~(\citeyear{liu2020incomplete}) & 47.36 & \textbf{49.59} \\
 \hline
 \multicolumn{4}{l}{\textbf{Dialog Rewrite}} \\
 C{\small ANARD}~(R-L) & 74.70~(\citeyear{liu2021topic}) & \textbf{83.91} & 83.87 \\
 \hline
\end{tabular}
}
\caption{The performance on generation tasks with more free and larger output space than knowledge distillation tasks.}
\vspace{-0.4cm}
\label{tab:gen}
\end{table}

\begin{table}[t]
\centering
\small
\newcolumntype{s}{>{\columncolor[HTML]{DCDCDC}} c}
\setlength{\tabcolsep}{0.8mm}{
\begin{tabular}{c c c c c} 
 \bottomrule
 \textbf{Model} & \textbf{Inform} & \textbf{Success} & \textbf{BLEU} & \textbf{Combined} \\ 
 \hline
 \multicolumn{5}{l}{\textbf{With Predicted Dialogue State}} \\
 SimpleTOD~(\citeyear{hosseini2020simple}) & 84.40 & 70.10 & 15.01 & 92.26 \\
 SOLOIST~(\citeyear{peng2021soloist}) & 85.50 & 72.90 & 16.54 & 95.74 \\
 MinTL~(\citeyear{lin2020mintl}) & 84.88 & 74.91 & 17.89 & 97.78 \\
 UBAR~(\citeyear{yang2021ubar}) & 91.50 & 77.40 & 17.00 & 101.50 \\
 NCM~(\citeyear{liu2021pretraining}) & 86.90 & 76.20 & 20.58 & 102.13 \\
 HTER~(\citeyear{santra2021hierarchical}) & 91.72 & 75.80 & 19.05 & 102.81 \\
 PPTOD~(\citeyear{su2021multi}) & 89.20 & 79.40 & 18.62 & 102.92 \\
 BORT~(\citeyear{sun2022bort}) & 93.80 & 85.80 & 18.50 & 108.30 \\
 GALAXY~(\citeyear{he2022galaxy}) & \textbf{94.40} & \textbf{85.30} & 20.50 & 110.35 \\
 DFM$_{\rm t5}$ & 91.50 & 84.70 & \textbf{22.86} & \textbf{110.96} \\
 \hline
 \multicolumn{5}{l}{\textbf{With Golden Dialogue State}} \\
 UBAR~(\citeyear{yang2021ubar}) & 94.00 & 83.60 & 17.20 & 106.00 \\
 GALAXY~(\citeyear{he2022galaxy}) & \textbf{94.80} & \textbf{85.70} & 19.93 & 110.18 \\
 DFM$_{\rm t5}$ & 93.20 & 85.60 & \textbf{23.38} & \textbf{112.78} \\
 \hline
\end{tabular}
}
\caption{The performance on end-to-end TOD M{\small ULTIWOZ2.0} benchmark.}
\label{tab:multiwoz_e2e}
\vspace{-3mm}
\end{table}

\subsection{Usage 3: Generation}
Different from structured semantic space in knowledge distillation, generation tasks support free and large output space, where our DFM also shows splendid efficiency. 
As shown in Table~\ref{tab:gen}, DFM$_{\rm t5}$ achieves comparable performance on dialogue generation tasks, which include two typical dialogue systems (Chitchat and QA). DFM$_{\rm t5}$ outperforms SOTA on three general generative tasks (chitchat, dialogue summary and dialogue rewrite). On QA task, the SOTA method leverages the data augmentation method, which is compatible with our method. In the future, we will try it on our model. To further validate the efficiency of DFM, we evaluate DFM$_{\rm t5}$ on well-studied end-to-end TOD task.

\noindent \textbf{Application in E{\small 2ETOD}} We conduct the experiments on M{\small ULTIWOZ2.0} benchmark. As shown in Table~\ref{tab:multiwoz_e2e}, there are four metrics: \textbf{Inform}, \textbf{Success}, \textbf{BLEU} and \textbf{Combined}. Inform and Success measure whether system response contains the user's constraints and matched entities. These two metrics measure the task completion ability of dialogue system. BLEU is BLEU-4 score, which measures the response consistency. \textbf{Combined} score equals to 0.5*(\textbf{Inform}+\textbf{Success})+\textbf{BLEU}. The dialogue state tracking is an indispensable part of TOD system, whose results are used to get database state. With predicted dialogue state, which is another DST model fine-tuned from DFM$_{\rm t5}$, our DFM$_{\rm t5}$ can achieve comparable results with SOTA methods. There is absolute 0.61 \textbf{Combined} score gain. However, with golden dialogue states, DFM$_{\rm t5}$ reaches over 112 \textbf{Combined} score.

\section{Conclusion}
In this paper, we collect 73 dialog datasets across 15 supervised dialog-oriented tasks (DialogZoo) to pre-train a unified dialog foundation model (DFM). 
To enhance dialogue-level representation ability of DFM, we further propose two unsupervised dialog denoising tasks to jointly pre-train the DFM with multitask learning. 
The experimental results show that DFM can obtain strong performance on three aspects: representation, knowledge distillation and generation. DFM contributes the first multi-purposed cross-domain pre-trained model with superior performance on numerous downstream tasks.



\bibliography{anthology,custom}
\bibliographystyle{acl_natbib}

\appendix


\section{Appendix}
\label{sec:appendix}

\begin{algorithm}[h!]
  Initialize the DFM $\Theta$ with backbone model. \\
  Set the max number of epoch: $E$. \\
  Task number of DOTs: $N$. \\
  Sample counts of DOTs: $C=\{c_i\}_{i=1}^N$. \\
  Initialize training step: $s=0$. \\
  \textit{// Prepare training data for DFM} \\
  \For{i in 1,\dots,N}{ 
    Reformulate dialogue-oriented samples of $i$-th task into text-to-text format as $D_i$.
  }
  \textit{// Start task-iterative training} \\
  \For{e in 1,\dots, E}{ 
    \While{not the final mini-batch of $\max{C}$}{
         \For{i in 1,\dots,N}{
            Sample a mini-batch in $i$-th task: $b_i$. \\
            Calculate loss: $L({\Theta})={\rm NLLLoss}(b_i)$. \\
            Compute gradient: $\nabla L({\Theta})$. \\
            Update: $\Theta \longleftarrow \Theta-\alpha \nabla L({\Theta})$
         }
    }
  }
  \caption{Task-iterative Training}
  \label{algo:tits}
\end{algorithm}

\begin{table*}
\centering
\small
\begin{tabular}{p{0.15\textwidth} | p{0.75\textwidth}}
\hline
\textbf{DOTs} & \textbf{Datasets} \\
\hline
R{\small{EW}} & Task~\cite{dctask}, Canard~\cite{dccanard}, Mudoco~\cite{mudoco} \\
N{\small{LG}} & Sclstm~\cite{sclstm}, E2E~\cite{e2e_dataset}, Rnnlg~\cite{rnnlg}, E2E-challenge~\cite{e2e_challenge}, Google-SIM~\cite{google-sim} \\
S{\small{UM}} & Samsum~\cite{samsum}, Dialogsum~\cite{dialogsum}, ReadingComprehension~\cite{readingcommprehension} \\
F{\small{ILL}} & Restaurant8k~\cite{restaurant8k}, Snips~\cite{snips}, HWU64~\cite{hwu64}, MAMS~\cite{MAMS}, ASTE~\cite{aste}, Sentihood~\cite{sentihood} \\
I{\small{NTENT}} & Banking77~\cite{banking77}, Snips~\cite{snips}, HWU64~\cite{hwu64}, Clinc~\cite{clinc} \\
D{\small{ST}} & SGD~\cite{rastogi2020towards}, TaskMaster2~\cite{tm2}, WOZ~\cite{woz}, MultiWOZ2.2~\cite{mwoz22}, MultiDogo~\cite{multidogo}, DSTC2~\cite{dstc2}, DSTC3~\cite{dstc3} \\
C{\small{OMM}} & Alphanli~\cite{alphanli}, Commonsense-qa~\cite{commonsenseqa}, Cosmosqa~\cite{cosmosqa}, Csqa2~\cite{csqa2}, SocialIQA~\cite{socialiqa} \\
E{\small{MO}} & Emory~\cite{emory}, Go-emotion~\cite{goemotion}, Meld~\cite{meld}, Reccon~\cite{reccon} \\
D{\small{OCQA}} & CoQA~\cite{coqa}, CMUDoG~\cite{cmudog}, DoQA~\cite{doqa}, NarrativeQA~\cite{narrativeqa}, QuAC~\cite{quac}, Race~\cite{race}, Squad~\cite{squad} \\
D{\small{IALQA}} & DDrel~\cite{ddrel}, FriendsQA~\cite{friendsqa}, Molweni~\cite{molweni}, DialogRE~\cite{dialogre}, Mutual~\cite{mutual}, Dream~\cite{dream} \\
C{\small{HAT}} & DailyDialog~\cite{dailydialog}, PersonaChat~\cite{personachat}, MetalWOZ~\cite{metalwoz}, EmpatheticDialog~\cite{empatheticdialog}, CommonsenseDialog~\cite{commonsensedialog} \\
K{\small{GDIAL}} & Soccer-kgdial, Incar-kgdial~\cite{soccerkgdial} \\
T{\small{XT2SQL}} & Spider~\cite{yu2018spider}, Sparc~\cite{yu2019sparc}, CoSQL~\cite{yu2019cosql} \\
S{\small{IM}} & SGD~\cite{rastogi2020towards}, Task-master2~\cite{tm2}, WOZ~\cite{woz}, MultiWOZ2.2~\cite{mwoz22}, MultiDogo~\cite{multidogo}, DSTC2~\cite{dstc2}, DSTC3~\cite{dstc3} \\
T{\small{OD}} & SGD~\cite{rastogi2020towards}, Task-master2~\cite{tm2}, WOZ~\cite{woz}, MultiWOZ2.2~\cite{mwoz22}, MultiDogo~\cite{multidogo}, DSTC2~\cite{dstc2}, DSTC3~\cite{dstc3} \\
\hline
\end{tabular}
\caption{\label{tab:datasets}
All the collected datasets for each dialogue-oriented tasks in DialogZoo.
}
\end{table*}

\clearpage
\onecolumn
\small
\begin{longtable}{p{0.07\textwidth}p{0.11\textwidth}p{0.8\textwidth}}
\toprule
\textbf{Task}        & \textbf{\begin{tabular}[c]{@{}c@{}}Source\\ Dataset\end{tabular}} & \multicolumn{1}{c}{\textbf{\begin{tabular}[c]{@{}c@{}}Example\\ (Input/Output)\end{tabular}}}                                                                                              \\ \midrule
\multirow{2}{*}{R{\small{EW}}} & \multirow{2}{*}{Canard}                                           & {[}rew{]} ernie kovacs {[}T{]} death {[}T{]} when did he die ? {[}EK{]} none {[}C{]} Semantically complete sentence 'when did he die ?'                                                \\
                     &                                                                   & {[}rew{]} when did ernie kovacs die ?                                                                                                                                                        \\ \midrule
\multirow{2}{*}{N{\small{LG}}} & \multirow{2}{*}{Google-Sim}                                       & {[}nlg{]} {[}EK{]} movies.notify\_success(theatre\_name=camera\_7; time=6:00\_pm; movie=a\_man\_called\_ove; date=march\_08) {[}C{]} With current system act, write a system response. \\
                     &                                                                   & {[}nlg{]} purchase is confirmed for march 08 at 6:00 pm to see a man called ove at camera 7                                                                                                  \\ \midrule
\multirow{2}{*}{S{\small{UM}}} & \multirow{2}{*}{Samsum}                                           & {[}sum{]} \#Person1\#: I am confused by what he said. {[}T{]} \#Person2\#: Why do you say that? {[}T{]} \#Person1\#: I don't know what he wants to do. Does he want help me or just scold me? {[}T{]} \#Person2\#: Think a little. I think he means well at the bottom of his heart. {[}EK{]} none  {[}C{]} Give a summary of this dialogue. \\
                     &                                                                   & {[}sum{]} \#Person1\# tells \#Person2\# \#Person1\#'s confused by the man's words.  \\ \midrule
\multirow{2}{*}{F{\small{ILL}}} & \multirow{2}{*}{Restaurant8k}                                           & [fill] Please book this hotel for 10 people and on Sun, 05 Aug 2018 [EK] restaurant.date  [C] Give the value of the slot. \\
                     &                                                                   & [fill] Sun, 05 Aug 2018  \\ \midrule
\multirow{2}{*}{I{\small{NTENT}}} & \multirow{2}{*}{HWU64}                                           & [intent] it's dirty here make some noise [EK] domain.home\_assistant [C] Give the user's intent on the domain. \\
                     &                                                                   & [intent] iot cleaning \\ \midrule
\multirow{2}{*}{D{\small{ST}}} & \multirow{2}{*}{SGD}                                           & [dst] USER: I need to find an apartment please. [T] SYSTEM: In what area? [T] USER: Something in San Jose. [EK] home.area [C] Give the user's constraint on the slot. \\
                     &                                                                   & [dst]  San Jose \\ \midrule
\multirow{2}{*}{C{\small{OMM}}} & \multirow{2}{*}{CSQA2}                                           & [comm] Consider the argument, is the argument correct? [EK] argument: Science has found a cure for cancer  [C] With given argument, answer the question. \\
                     &                                                                   & [comm] no \\ \midrule
\multirow{2}{*}{E{\small{MO}}} & \multirow{2}{*}{Emory}                                           & [emo] Marsha : Well, she has issues. [T] How does Marsha feel? [EK] possible choices: Joyful | Neutral | Powerful | Mad | Sad | Scared | Peaceful [C] With given possible emotions, select the correct answer. \\
                     &                                                                   & [emo] Sad \\ \midrule
\multirow{2}{*}{D{\small{OCQA}}} & \multirow{2}{*}{CoQA}                                           & [docqa] When was the Vat formally opened? [EK] The Vatican Apostolic Library (), more commonly called the Vatican Library or simply the Vat, is the library of the Holy See, located in Vatican City. Formally established in 1475, although it is much older, it is one of the oldest libraries in the world and contains one of the most significant collections of historical texts. It has 75,000 codices from throughout history, as well as 1.1 million printed books, which include some 8,500 incunabula.  $\cdots$  [C] With given background, context and previous questions, answer the question. \\
                     &                                                                   & [docqa] It was formally established in 1475 \\ \midrule
\multirow{2}{*}{D{\small{IALQA}}} & \multirow{2}{*}{Dream}                                           & [dialqa] W: Well, I'm afraid my cooking isn't to your taste. [T] M: Actually, I like it very much. [T] W: I'm glad you enjoy it. Let me serve you some more fish. [T] M: No, thank you. I've had enough fish, but I'd like some soup. [T] W: Here it is. Help yourself! [T] M: Thanks. I didn't know you were so good at cooking. If only my wife could learn to cook from you. [T] W: Why not bring your wife next time? I haven't seen her for quite a while. [T] M: OK, I will. She will be very glad to see you, too. Thank you for the wonderful meal. [T] What does the man think of the woman's cooking?? [EK] possible choices: It's really terrible. | It's very good indeed. | It's better than what he does. [C] With given dialogue and choices, select the correct answer. \\
                     &                                                                   & [dialqa] It's very good indeed. \\ \midrule
\multirow{2}{*}{C{\small{HAT}}} & \multirow{2}{*}{PersonaChat}                                           & [chat] other: Hi, how are you doing? I'm getting ready to do some cheetah chasing to stay in shape. [T] speaker : You must be very fast. Hunting is one of my favorite hobbies. [T] other: I am! For my hobby I like to do canning or some whittling. [EK] Speaker's personality is following: I like to remodel homes. | I like to go hunting. | I like to shoot a bow. | My favorite holiday is halloween. [C] With given speaker's personality and previous dialogue, give a speaker response. \\
                     &                                                                   & [chat] I also remodel homes when I am not out bow hunting. \\ \midrule
\multirow{2}{*}{K{\small{GDIAL}}} & \multirow{2}{*}{Incar-Kgdial}                                           & {[}kgdial{]} i m in need of coffee is there any near my current location {[}EK{]} starbucks.distance.4\_miles | starbucks.traffic\_info.moderate\_traffic | starbucks.poi\_type.coffee\_or\_tea\_place | starbucks.address.792\_bedoin\_street | home.distance.4\_miles | home.traffic\_info.moderate\_traffic | home.poi\_type.home | home.address.5671\_barringer\_street {[}C{]} Please answer the user's question accordding to the knowledge graph. \\
                     &                                                                   &  {[}kgdial{]} starbucks and teavana are close \\ \midrule
\multirow{2}{*}{T{\small{XT2SQL}}} & \multirow{2}{*}{CoSQL}                                           & [txt2sql] Find out the average salary of professors? [T] Find the average salary of the professors of each department? [EK] * | classroom.building | classroom.room\_number | classroom.capacity | department.dept\_name | department.building | department.budget | course.course\_id | course.title | course.dept\_name | course.credits | instructor.ID | instructor.name | instructor.dept\_name | instructor.salary | section.course\_id | section.sec\_id | section.semester | section.year | section.building | section.room\_number | section.time\_slot\_id | teaches.ID | teaches.course\_id | teaches.sec\_id | teaches.semester | teaches.year | student.ID | student.name | student.dept\_name | student.tot\_cred | takes.ID | takes.course\_id | takes.sec\_id | takes.semester | takes.year | takes.grade | advisor.s\_ID | advisor.i\_ID | time\_slot.time\_slot\_id | time\_slot.day | time\_slot.start\_hr | time\_slot.start\_min | time\_slot.end\_hr | time\_slot.end\_min | prereq.course\_id | prereq.prereq\_id [C] Translate the user's question into SQL query.
 \\
                     &                                                                   & [txt2sql] SELECT AVG(instructor.salary) , instructor.dept\_name FROM instructor GROUP BY instructor.dept\_name
 \\ \midrule
\multirow{2}{*}{S{\small{IM}}} & \multirow{2}{*}{WOZ}                                           & {[}sim{]} USER: Hi, I'm looking for a nice German restaurant. {[}T{]} SYSTEM: There are no German restaurants listed, do you have another preference? {[}EK{]} final dialogue state: (restaurant\_search.food.european | restaurant\_search.area.dontcare | restaurant\_search.price\_range.expensive | restaurant\_search.request.phone | restaurant\_search.request.address | restaurant\_search.request.area) {[}C{]} With given dialogue history and final dialogue state, give a user utterance. \\
               &                                                                   & [sim] How about European yeah? \\ \midrule
\multirow{2}{*}{T{\small{OD}}} & \multirow{2}{*}{TaskMaster2}                                           & [tod] SYSTEM: Hello user. How can I help you? [T] USER: Can you play music for me? [T] SYSTEM: Can you please name the artist? [T] USER: Can you play something by Whitney Houston. [EK] current dialogue state: (music.name\_artist.Whitney\_Houston) [C] With given dialogue history and current dialogue state, write a system response. \\
               &                                                                   & [tod] Whitney Houston, let me find you something. \\ \midrule
\multirow{2}{*}{C{\small{LO}}} & \multirow{2}{*}{DialogZoo}                                           & [clo] USER: hi,good <mask> [T] USER: I need a copy of insurance for my car [T] SYSTEM: Hey there!  How may I help you <mask>? [T] SYSTEM: Absolutely! I'll get that done for you. May I know your <mask> and last name and the <mask> digits of your <mask> number please? [T] USER: name: sofia and ssn no: <mask> [EK] 3468 | first | last 4 | morning | today | SSN [C] Please cloze the dialogue content.
 \\
               &                                                                   & [clo] USER: hi,good morning [T] USER: I need a copy of insurance for my car [T] SYSTEM: Hey there!  How may I help you today? [T] SYSTEM: Absolutely! I'll get that done for you. May I know your first and last name and the last 4 digits of your SSN number please? [T] USER: name: sofia and ssn no: 3468
 \\  \midrule
\multirow{2}{*}{R{\small{EO}}} & \multirow{2}{*}{DialogZoo}                                           & [reo] Find the physicians who prescribe this medication. [T] Which medication has brand X? [T] What are their name and position? [EK] none [C] Please reorder the dialogue content.
 \\
               &                                                                   & [reo] Which medication has brand X? [T] Find the physicians who prescribe this medication. [T] What are their name and position?
 \\  
\bottomrule
\caption{Reformulated text-to-text examples of all the dialogue-oriented tasks used to train the dialogue foundation models.} \\
\end{longtable}

\end{document}